\documentclass[Journal,letterpaper]{ascelike-new}
\usepackage[utf8]{inputenc}
\usepackage[T1]{fontenc}
\usepackage{lmodern}
\usepackage{graphicx}
\usepackage[figurename=Fig.,labelfont=bf,labelsep=period]{caption}
\usepackage{subcaption}
\usepackage{amsmath}
\usepackage{newtxtext,newtxmath}
\usepackage[colorlinks=true,citecolor=red,linkcolor=black]{hyperref}
\NameTag{Zhu, \today}
\begin{document}
\nolinenumbers

\title{Intelligent Traffic Light via Policy-based Deep Reinforcement Learning}
\author[1]{Yue Zhu}
\author[2] {Mingyu Cai}
\author[3]{Chris Schwarz}
\author[1] {Junchao Li}
\author[4]{Shaoping Xiao}

\affil[1]{Department of Mechanical Engineering, The University of Iowa, Iowa City, IA 52242}
\affil[2]{Department of Mechanical Engineering, Leigh University, Bethlehem, PA, 18015}
\affil[3]{National Advanced Driving Simulator, The University of Iowa, Iowa City, IA 52242}
\affil[4]{3131 Seamans Center, Department of Mechanical Engineering, Iowa Technology Institute, The University of Iowa, Iowa City, IA 52242. Email: shaoping-xiao@uiowa.edu}

\maketitle

\begin{abstract}
Intelligent traffic lights in smart cities can optimally reduce traffic congestion. In this study, we employ reinforcement learning to train the control agent of a traffic light on a simulator of urban mobility. As a difference from existing works, a policy-based deep reinforcement learning method, Proximal Policy Optimization (PPO), is utilized other than value-based methods such as Deep Q Network (DQN) and Double DQN  (DDQN). At first, the obtained optimal policy from PPO is compared to those from DQN and DDQN. It is found that the policy from PPO performs better than the others. Next, instead of the fixed-interval traffic light phases, we adopt the light phases with variable time intervals, which result in a better policy to pass the traffic flow. Then, the effects of environment and action disturbances are studied to demonstrate the learning-based controller is robust. At last, we consider unbalanced traffic flows and find that an intelligent traffic light can perform moderately well for the unbalanced traffic scenarios, although it learns the optimal policy from the balanced traffic scenarios only.
\end{abstract}

\section{Introduction}

As the foundation of our society, transportation systems help ensure that people can reach every destination. Furthermore, transportation promotes economic growth via increasing business productivity, enhancing accessibility of labor force and jobs, and improving supply chain efficiency. However, traffic congestion has become more and more costly. According to data analyzed by INRIX in \citeyearNP{INRIX2018}, traffic congestion has cost each American 97 hours and $ \$1,348$ per year. In addition to the waste of fuel, traffic congestion increases carbon emissions \cite{Zhang2013,Bharadwaj2017} as one of the most harmful effects on the environment. Since traffic lights have been used to control traffic flow, one of the solutions to mitigating traffic congestion is maximizing the traffic light performance by an optimal control strategy. 
 
The earliest traditional traffic light control approach includes predefined fixed-time plans \cite{Miller1963}, in which formulas were derived to predict the average delay to vehicles with the consideration of fixed-cycle traffic lights. In another approach, \citeN{Cools2013} implemented self-organizing traffic lights via actuated control in an advanced traffic simulator with real data. On the other hand, \citeN{Zhou2011} investigated adaptive traffic light control of multiple intersections using real-time traffic data. The results demonstrated that the adaptive control could produce lower average waiting time and fewer stops than the predefined fixed-time control and the actuated control. In addition, another adaptive traffic light control \cite{Miao2021} was proposed for connected and automated vehicles at isolated traffic intersections. This control approach not only reduces the average waiting time but also guarantees the worst-case waiting time. Furthermore, \citeN{Demitrov2020} developed a method to improve the level of traffic light service by optimizing the phase length and cycle.

Riding the wave of artificial intelligence (AI), deep learning (DL) and reinforcement learning (RL) have been employed in solving various engineering problems \cite{Xiao2020,Cai2021}. As a subset of machine learning, RL \cite{Sutton2018} enhances the control agent to obtain an optimal action strategy, i.e., policy, during the interaction with the environment. There is a growth of interest in learning-based control of traffic lights \cite{Bingham2001,Kuyer2008}, i.e., intelligent traffic lights. \citeN{Li2016} proposed algorithms to design traffic signal timing plans by setting up a deep neural network (DNN) to learn the state-action value function (also called Q-function) of RL. In this deep Q-learning approach, the agent learned appropriate signal timing policies from the sampled traffic state, control inputs, and the corresponding traffic system performance output. \citeN{Wei2018} also employed deep Q-learning to train the traffic light control agent. They tested the method on a large-scale real traffic dataset obtained from surveillance cameras. 

In addition, multi-agent reinforcement learning (MARL) was employed to coordinate the traffic light controllers of multiple intersections. \citeN{Wu2020} proposed a novel algorithm for traffic light control in vehicular networks. They considered both the vehicles and the pedestrians who waited to pass through the intersection. The experimental results showed that their method could run stably in various scenarios. \citeN{Wang2020} designed a graph neural network-based model to represent interactions between multiple traffic lights. Then, a deep Q-learning method was utilized to make operation decisions for each traffic light. \citeN{Chen2020} used a concept of "pressure" in RL so that the designed control agents could coordinate multiple traffic lights. They experimented on a real-world scenario with more than two thousand traffic lights in Manhattan, New York City. 

In this study, we utilize a deep RL method to synthesize an optimal operation strategy of a traffic light to pass the traffic flow intelligently. The contributions are manifold, as described below. Most existing works employed value-based RL methods, such as deep Q-learning \cite{Wei2018,Wang2020,Chen2020}, to train the control agent. As a difference, we adopt a policy-based RL method, Proximal Policy Optimization (PPO), in this work and compare the results with the ones obtained from value-based RL methods. On the other hand, some previous works considered only two traffic light phases \cite{Li2016,Wei2018} or four phases \cite{Wu2020} at the intersection. Although only one traffic intersection is studied in this work, we investigate a complex traffic system, which allows left-turn, right turn, and U-turn in each branch. Therefore, a total of eight traffic light phases are considered. Furthermore, we consider the traffic light with variable-interval phases. As a result, it performs better than the one with fixed-interval phases that have been used in most previous studies \cite{Wei2018,Wei2019}, mainly resulting in fewer stops. In addition, two traffic scenarios are studied to demonstrate the robustness of the intelligent traffic light trained via PPO: (1) environment disturbance because of car accidents and (2) action disturbance due to traffic light malfunction. According to our best knowledge, none of them has been investigated in the literature. At last, we find that an intelligent traffic light can operate moderately well under unbalanced traffic flows, although it only learns from the balanced traffic scenarios.

The paper is organized as follows. Section 2 introduces deep RL methods, including value-based and policy-based methods, for the control of traffic lights. Section 3 compares the optimal policies obtained from various RL methods. Then, the advantages of the traffic light phases with variable time intervals in passing the traffic flow are investigated. Section 3 also considers environment and action disturbances to demonstrate the robustness of the optimal policies. Finally, Section 4 provides conclusions and future works.

\section{Methodology}

\subsection{Reinforcement learning problem of traffic light control}

\begin{figure}
\centering{}
\includegraphics[scale=0.7]{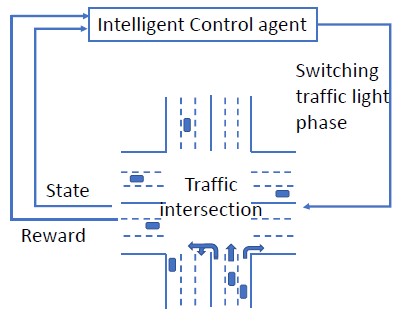}
\caption{The interaction between the control agent and the traffic intersection in reinforcement learning.}
\label{fig:RL}
\end{figure}

When formulating an RL problem in this study, the control of a traffic light can be described by the traffic light, as a controller or a control agent, interacting with the environment, i.e., a traffic intersection, as shown in Fig.~\ref{fig:RL}. The learning process is iterative. At each iteration, the control agent observes the state of the traffic intersection and decides an action - switching to a new traffic light phase or staying with the current phase. The traffic flow at the intersection will be correspondingly changed due to the decision-making of the control agent. When observing the next state of the traffic intersection, the control agent receives feedback called reward and then decides the following action. 

In addition to an agent, key components in basic RL include an environment, states, actions, and a reward function. In this study, we consider a four-way traffic intersection, where there are three 300-meter-long lanes in each incoming or outgoing road. The outside lane is for right turn only, while the middle lane is for straight going, i.e., through movement. The inside lane allows both left-turn and U-turn, shown in Fig.~\ref{fig:RL}. This traffic intersection is modeled and simulated via an open-source and highly portable traffic simulator called Simulation of Urban Mobility (SUMO) \cite{SUMO2018}. It is assumed that all vehicles are the same type with the information in Table~\ref{table:vehicledata}. The car length is defined as the distance from the front bumper to the rear bumper, while the min gap means the distance between the rear car's rear bumper and the rear car's front bumper when stopping. SUMO randomly inputs the vehicles at the end of each incoming road, following the traffic flow setting listed in Table~\ref{table:flowrate}, until reaching the desired number of vehicles that is 808 in this study. 

\begin{table}
\caption{Vehicle data in Simulation of Urban Mobility (SUMO)}
\label{table:vehicledata}
\centering
\small
\renewcommand{\arraystretch}{1.25}
\begin{tabular}{l l}
\hline\hline
\multicolumn{1}{c} {Attribute} &
\multicolumn{1}{c}{Values} \\
\hline
Length ($m$) & 3 \\
Min gap ($m$) & 2  \\
Max acceleration ($m/s^2$) & 1 \\
Max deceleration($m/s^2$) & 4.5 \\
\hline\hline
\end{tabular}
\normalsize
\end{table}

\begin{table}
\caption{Traffic flow rates}
\label{table:flowrate}
\centering
\small
\renewcommand{\arraystretch}{1.25}
\begin{tabular}{l l}
\hline\hline
\multicolumn{1}{c} {Vehicle route} &
\multicolumn{1}{c}{Rate (vehicles per hour)} \\
\hline
Right-turn & 480 \\
Through movement & 600  \\
Left-turn & 240 \\
U-turn & 120 \\
\hline\hline
\end{tabular}
\normalsize
\end{table}

It is assumed that the environment is fully observable so that the control agent has complete knowledge about the traffic status when observing the intersection. Such a so-called state is represented by vehicles' positions, velocities, and waiting times as well as the current traffic light phase. All information can be withdrawn from the SUMO simulator. Specifically, each road is discretized into many cells \cite{Liang2019}, and every cell has the same width as the lane width and the length as the summation of the car length and a min gap. If a car’s center is located in a cell, 1 is assigned to this cell. Otherwise, there is 0 assigned to the cell. Consequently, all cells form a vehicle position matrix in which the number of rows equals the number of roads, and the number of columns is the number of cells on each road. Similarly, the vehicle velocity and waiting-time matrices can be generated by assigning the vehicles' velocities and waiting times in the corresponding cells. In addition, the traffic light phase is encoded as a vector of 1's (green lights) and 0's (red lights). In summary, the state variables include the vehicle position, velocity, and waiting-time matrices and the traffic light phase vector.

\begin{figure}
\centering{}
\includegraphics[scale=0.7]{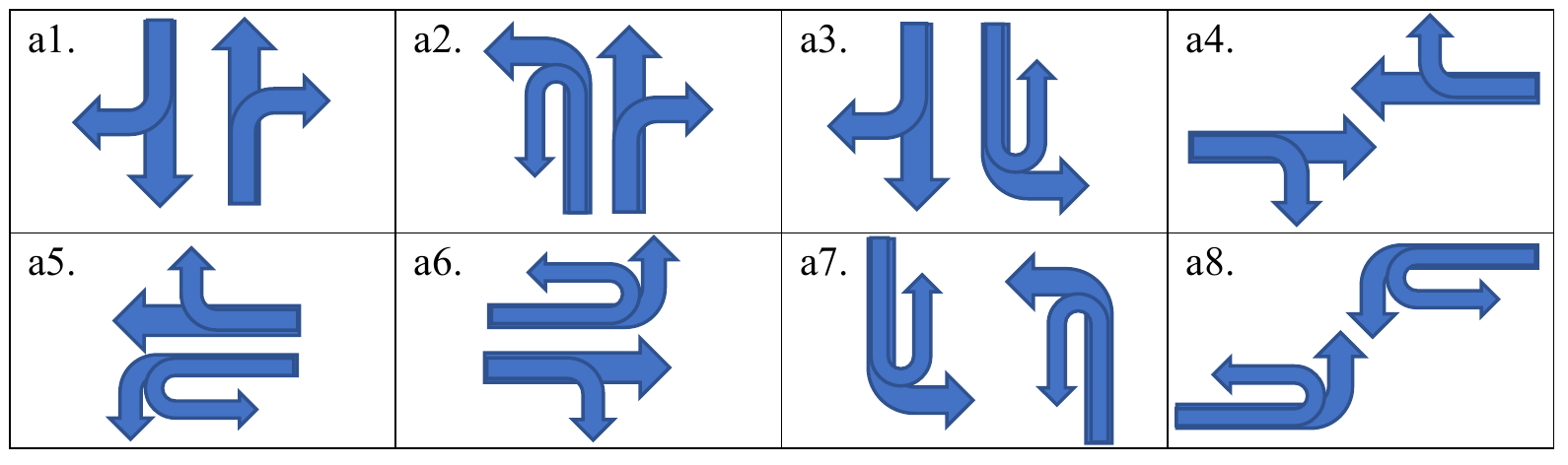}
\caption{Traffic light phases.}
\label{fig:actions}
\end{figure}

The agent can take eight actions to switch the current traffic light phase to one of the phases in Fig.~\ref{fig:actions}. We first consider the traffic light phases with fixed time intervals. There is a 5-second transition if the agent chooses a traffic light phase different from the current one, and the chosen light phase will last 10 seconds. Otherwise, the current light phase will be extended to five more seconds. It shall be noted that later in this study, we will consider the traffic light phases with variable time intervals. 

The reward function defines the goal of an RL problem. It gives feedback about good or bad events after the agent takes the selected action and reaches the next state. The agent's objective is to maximize the accumulative reward over the long run. In this study, the reward function is related to traffic light phase switching and traffic status at the interaction, as expressed below.
\begin{equation} \label{eq:reward}
R = R_a - R_1 - 0.5 R_2 + 0.8 R_3
\end{equation}
where $R_a$ is the action reward. If the agent chooses a different light phase, $R_a=-5$; otherwise, $R_a=0$. After the agent reaches the next state, the other reward components in Eq.~\ref{eq:reward} can be calculated based on the vehicle position, velocity, and waiting-time matrices. $R_1$ is the total number of vehicles that are stopped on all four roads at the next state. A vehicle is considered as stopped if its velocity is below 0.1 $m/s$. $R_2$ is the average waiting time (in seconds) of all stopped vehicles by the time of the next state. Once a vehicle starts to move, its waiting time is reset as zero. 

The last reward component, $R_3$, in Eq.~\ref{eq:reward} is calculated as
\begin{equation} \label{eq:rewardR3}
R_3 = \sum_{i=1}^{N_l} R^l_i = \sum_{i=1}^{N_l} 0.02 (n_{avg} - n_i) n_i
\end{equation}
where $N_l$ is the total number of lanes at the traffic intersection, $n_{avg}=R_1/{N_l}$ is the average number of stopped vehicles, and $n_i$ is the number of stopped vehicles at lane $i$. It shall be noted that $R_a$, $R_1$, and $R_2$ are commonly used in existing works \cite{Nishi2018}. The other widely-used reward components in studies of intelligent traffic lights include the total length of waiting vehicles and the number of vehicles that have passed the traffic light \cite{Wei2018}. Indeed, $R_1$ globally quantifies the traffic congestion at the intersection while $R_3$ represents how well the local traffic congestion is balanced at individual lanes. Per the authors' knowledge, this paper is the first to introduce the balance of traffic congestion in the reward to train an intelligent traffic light. 

\subsection{Q-learning and value-based methods}

By maximizing the accumulated reward, the control agent will find the optimal policy to reduce the traffic congestion as much as possible. A policy defines decision-making, i.e., the action selection, by the agent at a given state. The optimal value function (either state value or state-action value) in value-based RL methods is directly solved. A value function specifies how well the goal is achieved by the RL agent over the long run. For example, a state-action value, which is also called Q value as $Q(s,a)$, is the total reward that an agent can expect to accumulate over a long run, starting from state $s$ and taking action $a$. Once the optimal value function is found, the optimal policy can be determined via the greedy action selection. 

Q-learning \cite{Watkins1992} is one of the value-based RL methods, and it is model-free because the transition function of state-to-state is not required. This method evaluates all actions at each state to determine the best move via Monte Carlo simulations. The na\"ive Q-learning method is a tabular method in which Q values at each state are stored in a so-called Q-table. This table can guide the agent to the best action with the highest Q value at each state. In each episode, the Q value at the current state $s$ when taking action $a$ is updated at every step as below based on the Bellman equation \cite{Sutton2018}. 
\begin{equation} \label{eq:Q-learning}
Q_{\text{new}}(s,a) = Q(s,a)+\alpha \left[ R(s,a)+\gamma \max_{a'} Q(s',a')-Q(s,a) \right]
\end{equation}
where $\max_{a'} Q(s',a')$ outputs the highest Q value at the next state $s'$. $\gamma$ is the discount factor, i.e., a number between 0 and 1, so that the total reward remains bounded. It also implies how important future rewards are. $\alpha$ is the learning rate. A large learning rate may have the Q values converge faster. However, the convergence sometimes may be unstable or reach the value function other than the optimal one. On the contrary, a small learning rate can make the converge procedure smoother and more stable, but it converges slower. In practice, a large learning rate is used at the beginning and then decreases with iterations, and it is called an adaptive learning rate. 

Given enough episodes in Q-learning, when the optimal value function $Q^*(s,a)$ is converged, the optimal policy $\pi^*(a|s)$ can be determined by
\begin{equation} \label{eq:optimalPolicy}
\pi^*(a|s) = \underset{a}{\mathrm{argmax}} \left[ Q^*(s,a) \right]
\end{equation}

Deep neural networks are always employed in RL, named deep reinforcement learning (DRL), to solve the problems in which the state space and/or the action space are enormously large, or infinite (i.e., continuous), such as the traffic state space in this study. Therefore,  most existing works \cite{Nishi2018,Li2016,Wei2018} in learning-based traffic light control employed value-based DRL methods, including Deep Q Network (DQN) \cite{MnihKSGAWR13} and Double Deep Q Network (DDQN) \cite{Hasselt2016}, which are extensions of Q-learning. Since the state space is continuous, it is impossible to utilize the tabular approach to store and withdraw Q values. Unlike the na\"ive Q-learning that uses a Q-table, DQN and DDQN employ artificial neural networks, called Q-networks, to map states to Q values. 

\begin{figure}
\centering{}
\includegraphics[scale=0.4]{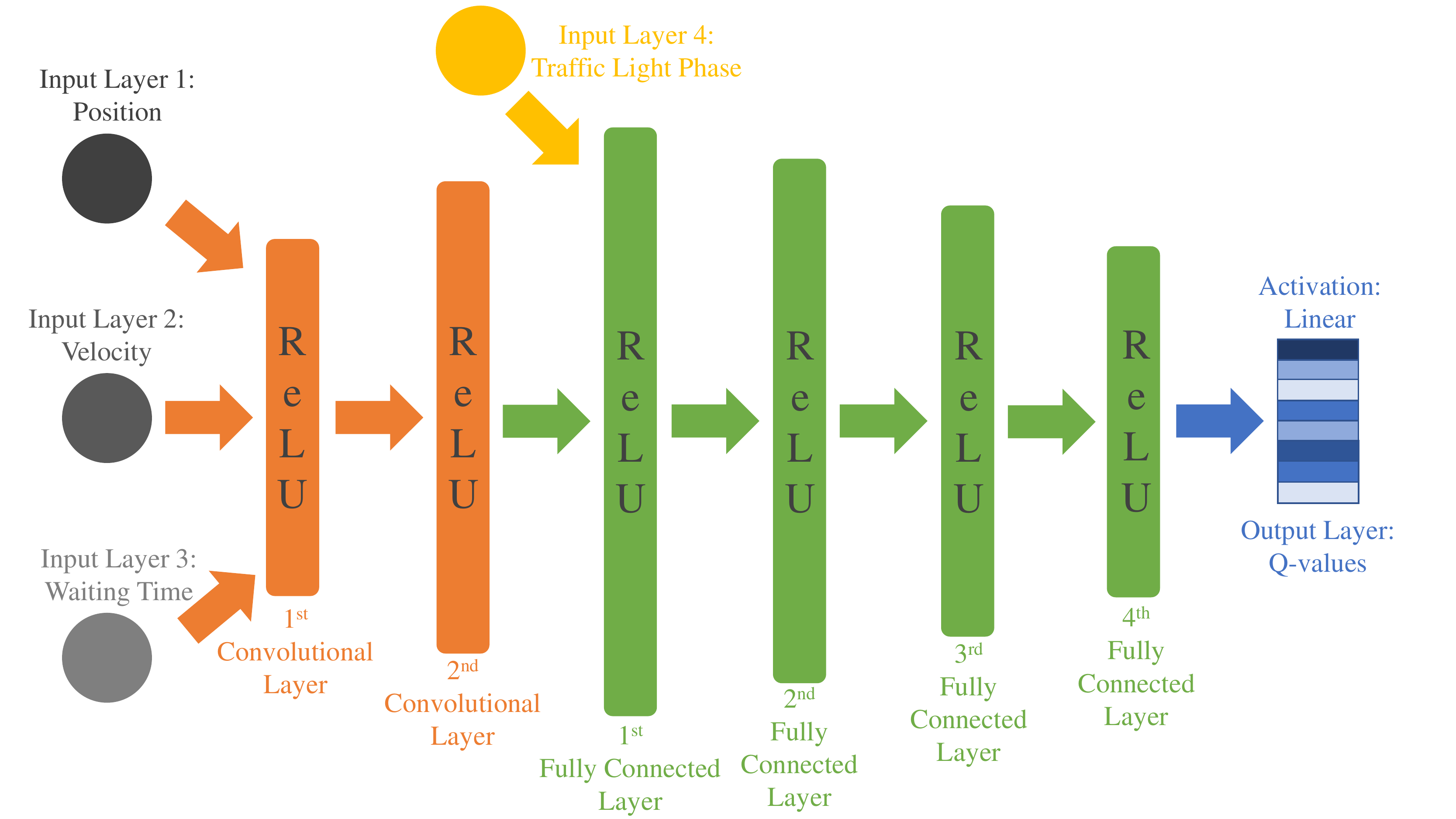}
\caption{The architecture of Q-networks.}
\label{fig:Qnets}
\end{figure}

DQN has two Q-networks, an evaluation Q-network and a target Q-network, which have the same architecture, as shown in Fig.~\ref{fig:Qnets}. The input features, including vehicle position, velocity, and waiting-time matrices, are processed via a convolutional neural network \cite{Lecun1998} with two layers. The first layer uses a kernel size of 4 by 4, a stride length of 2, and 16 out channels. The second layer uses a kernel size of 2 by 2, a stride length of 1, and 32 out channels. The rectified linear unit (ReLU) \cite{Nair2010} is used as the activation function. The output matrix and the traffic light phase vector are then flattened and passed to a fully-connected neural network to predict Q values. The fully-connected neural network has four hidden layers with 512, 256, 128, and 64 neurons. Since there are eight traffic light phases the control agent can choose, the output layer has eight neurons to predict Q values for individual actions. ReLu activation function is also used in the fully-connected neural network except for the output layer in which there is a linear activation function.   

During the learning process, the evaluation Q-network is updated each step by randomly selecting a batch of data samples. At the same time, the target Q-network holds fixed weights until copying from the evaluation Q-network once in a while (every 50 steps in this study). At each step, the evaluation Q-network predicts Q values at the current state so that the agent can choose an action $a$ at this state via the $\epsilon-$greedy technique \cite{Sutton2018}. After reaching the next state $s'$, the Q-networks are used to calculate the Q value when taking action $a$ at state $s$ via a revision of Eq.~\ref{eq:Q-learning} as
\begin{equation} \label{eq:Deep Q-learning}
Q_{\text{new}}(s,a) = Q_e(s,a)+\alpha \left[ R(s,a)+\gamma \max_{a'} Q_t(s',a')-Q_e(s,a) \right]
\end{equation}
where $Q_e$ represents the Q values predicted from the evaluation Q-network, and $Q_t$ represents the Q values predicted from the target Q-network. Eq.~\ref{eq:Deep Q-learning} generates one data sample, i.e., one experience, at each step. As an off-policy RL method, DQN also adopts experience replay memory \cite{Lin1992} that makes learning efficient. When applying DQN in this study, a batch of 32 experiences is randomly selected from the experience memory each step to update the evaluation Q-network. 

However, DQN sometimes overestimates Q values. Thus, DDQN \cite{Hasselt2016} is proposed by modifying how to update Q values after the action $a$ is taken at the current state $s$ and the next state $s'$ is reached. Recall that DQN directly uses the maximum Q-value at the next state $s'$ from the target Q-network on the right-hand side of Eq.~\ref{eq:Deep Q-learning} to update the Q value of action $a$ at the current state $s$. As a difference, shown in Eq.~\ref{eq:Double Deep Q-learning}, DDQN selects the action with the highest Q value at the next state $s'$ from the evaluation Q-network and then uses the Q value of the selected action at state $s'$ from the target Q-network to update the Q value of action $a$ at state $s$.
\begin{equation} \label{eq:Double Deep Q-learning}
Q_{\text{new}}(s,a) = Q_e(s,a)+\alpha \left[ R(s,a)+\gamma Q_t(s',\underset{a'}{\mathrm{argmax}} Q_e(s',a'))-Q_e(s,a) \right]
\end{equation}

\subsection{Policy-based reinforcement learning}

Unlike the value-based RL method, policy-based RL methods directly update and converge the optimal policy. Usually, they have good convergence properties and can learn stochastic policies, defined as $\pi_{\theta}(a|s)$ with a vector of policy parameters $\theta$, representing the probability of action $a$ to be chosen at state $s$. A commonly used loss function in policy gradient methods \cite{Kakade2002}  is empirically averaged over a finite batch of experiences
\begin{equation} \label{eq:Lloss}
\mathcal{L}(\theta)= \mathop{\mathbb{E}_t} \left[ \log \pi_{\theta} (a_t | s_t) \hat{A}_t \right]
\end{equation}
where $\hat{A}_t$ is an estimator of the advantage function at time $t$. Differentiating the objective function, i.e., the loss function, in Eq.~\ref{eq:Lloss} results in a gradient estimator \cite{Mniha16,schulman2018} during the optimization procedure to update policy parameters. However, the advantage function estimate $\hat{A}_t$ is usually very noisy, leading to destructively large policy updates.

\citeN{Schulman2015} proposed a trust region policy optimization (TRPO), in which a surrogate objective is maximized subject to a constraint on the limit of policy updates. Such an optimization problem is expressed as
\begin{equation} \label{eq:TRPO}
\max_{\theta} \mathop{\mathbb{E}_t} \left[ \frac{\pi_{\theta} (a_t | s_t)}{\pi_{\theta_{old}} (a_t | s_t)} \hat{A}_t \right]
\quad \text{subject to} \quad \mathop{\mathbb{E}_t} \left[ KL[\pi_{\theta_{old}}(\cdot | s_t), \pi_{\theta}(\cdot | s_t)] \right] \leq \delta
\end{equation}
where $\theta_{old}$ is the vector of policy parameters before the update, and $KL$ represents Kullback-Leibler divergence to measure the relative difference between the current and the old policies at a given state $s_t$. 

While keeping the benefits of TRPO, a new family of policy gradient methods, called Proximal policy optimization (PPO) \cite{schulman2017PPO}, revise the objective function in Eq.~\ref{eq:TRPO} as unconstrained optimization problems so that they are easy to implement and have a good sample complexity. One approach uses a penalty on the KL divergence with the adaptive penalty coefficient to achieve a target value of the KL divergence. In this study, we adopt another approach, developing a clipped surrogate objective, which performs better than the objective function with KL penalty. 

Let $r_t(\theta) = \frac{\pi_{\theta} (a_t | s_t)}{\pi_{\theta_{old}} (a_t | s_t)}$ denote the probability ratio of the current policy and the old policy. The clipped surrogate objective in PPO can be written as
\begin{equation} \label{eq:LPPO}
\mathcal{L}^{CLIP}(\theta)= \mathop{\mathbb{E}_t} \left[ \min(r_t (\theta)) \hat{A}_t, clip \left( r_t(\theta), 1-\epsilon, 1+\epsilon \right) \hat{A}_t \right]
\end{equation}
where $ clip \left( r_t(\theta), 1-\epsilon, 1+\epsilon \right) \hat{A}_t $ removes the incentive for moving $r_t$ outside of the interval $[1-\epsilon, 1+\epsilon]$, i.e., clipping the probability ratio to modify the surrogate objective. Such a way results in a lower bound on the original surrogate objective and guarantees the objective improvement. We use $\epsilon = $ 0.15 as the clipping range in this study.

PPO can be used in problems with either continuous or discrete action space. In this research, the action space is finite, i.e., discrete, as described above. Three artificial neural networks are utilized in our PPO methods, including actor-new network, actor-old network, and critic network. All the networks have the same architecture as Q-networks in Fig.~\ref{fig:Qnets} and take the state variables (the vehicle position/velocity/waiting-time matrices and the traffic light phase vector) as the inputs. However, they have different output layers. The actor networks output the probability of actions at a given state. Therefore, although there are still eight neurons on the output layer of actor networks as in Q-networks, the Softmax activation function \cite{Goodfellow2016} is employed other than the linear activation function. The critic network outputs the state value for advantage function estimation \cite{Schulman2018GAE} so that there is only one neuron with a linear activation function on the output layer.

This method is an on-policy learning method and only uses every collected experience once. Consequently, deep neural networks are updated once enough data samples are collected, and all old experiences will be discarded after updating. Here describes the network updating procedure. At each step, the current state is the input into the actor-new network to predict the probabilities of actions with which the agent can select the move. After taking the selected action, the agent reaches the next state and receives a reward. The current state, action, and reward are stored as one experience. This process is iterated for a certain number of steps until enough experiences are collected. This study uses a batch size of 100 to train and update the actor-new and critic networks. The actor-old network is a copy of the actor-new network before updating, and it represents the old policy in the probability ratio function ($r_t(\theta)$). During updating the actor-new network, the actor-old network always remains the same. 

\section{Simulations and Discussions}
In this study, Monte Carlo simulations with up to 600 episodes are conducted to train the control agent of traffic lights via RL. Each episode terminates when all vehicles pass the traffic interaction. All simulations are performed on a High-Performance Computing (HPC) cluster that is equipped with a GPU of 2080ti and a CPU with a frequency of 2.1 GHz. We first compare the performances of control agents trained via three RL methods: DQN, DDQN, and PPO to show that PPO is better than the other two methods. Then, the traffic light phases with variable time intervals are implemented, and the induced optimal policy is assessed. Furthermore, environment and action disturbances are studied to investigate the robustness of learning-based traffic lights. The above studies mainly consider balanced traffic flows that correspond to the SUMO setting described in Table 2. Finally, we investigate the scenarios of unbalanced traffic flows. Some simulation videos are provided \footnote{\url{https://github.com/YueZhu95/Intelligent-Traffic-Light-via-Reinforcement-Learning}} to demonstrate the operations of intelligent traffic lights under induced optimal policies. 

\subsection{Intelligent traffic lights trained via different DRL methods}

\begin{figure}
\centering{}
\includegraphics[scale=0.7]{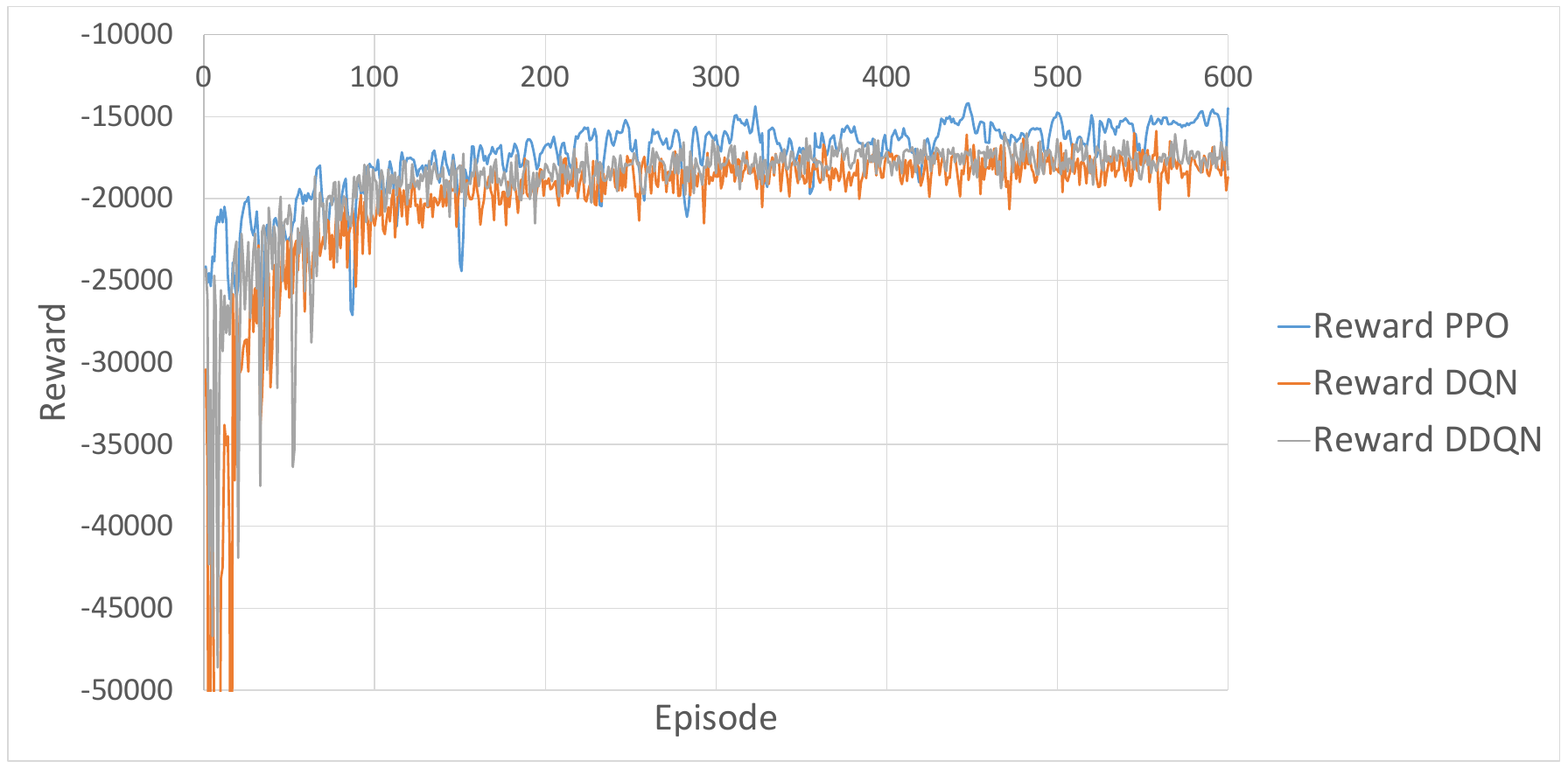}
\caption{Reward evolution during the training via DQN, DDQN, and PPO.}
\label{fig:reward3RL}
\end{figure}

Solving an RL problem aims to achieve an optimal policy, maximizing the accumulated reward over the long run. To compare the training performances of DQN, DDQN, and PPO, the collected reward as a function of episodes is illustrated in Fig.~\ref{fig:reward3RL}. The reward evolution represents the convergence of the training process. Once the training is converged, the accumulated reward can be one of the metrics to assess the performances of various RL methods if the problem setting keeps the same. A higher reward means that the induced policy is better. According to Fig.~\ref{fig:reward3RL}, DQN and DDQN have similar rewards after convergence, and PPO results in a higher reward than the others. In other words, PPO is better than DQN and DDQN in training intelligent control agents in this study. 

To assess the optimal policies obtained from three different DRL methods, we conducted ten simulations under each policy, respectively, to average the accumulated reward, the time with which all vehicles pass the traffic intersection, and the vehicles' total waiting time. The setting of SUMO environment keeps the same as the one used in learning. Additionally, we also collected the vehicles' passing time and waiting time from a predefined traffic light as a baseline. Finally, the predefined traffic light is set to repeatedly loop the traffic light phase from the first phase to the last one, as defined in Fig.~\ref{fig:actions}. The comparison is shown in Table~\ref{table:policyAssess}. 

It can be seen that the policies induced from DQN and DDQN result in almost the same reward, being consistent with the observation from Figure.~\ref{fig:reward3RL}. Also, when the intelligent traffic light operates under either the DQN- or DDQN-induced policies, it takes about 1250 seconds to pass all vehicles, i.e., 30$\%$ time reduction compared to the predefined traffic light. Considering the total of vehicles' waiting time, the DQN- and DDQN-induced policies result in 41$\%$ and 43$\%$ less time, respectively, than the predefined traffic light control strategy. In addition, the same conclusion that the PPO-induced policy is better than the others can be withdrawn from Table~\ref{table:policyAssess} as the one from Fig.~\ref{fig:reward3RL}. Therefore, under the PPO-induced policy, the intelligent traffic light can be more efficient to reduce traffic congestion than those under the DQN- and DDQN-induced policies. Specifically, the vehicles' passing time is reduced by 34$\%$, and the vehicles' total waiting time is reduced by 55$\%$, compared to the times resulting from the predefined traffic light.

\begin{table}
\caption{Assessment data of predefined traffic light and the learning-based traffic light via DQN, DDQN, and PPO}
\label{table:policyAssess}
\centering
\small
\renewcommand{\arraystretch}{1.25}
\begin{tabular}{c c c c}
\hline\hline
\multicolumn{1}{c} {Method} & 
\multicolumn{1}{c}{Reward}  & 
\multicolumn{1}{c} {Vehicle passing time (s)} & 
\multicolumn{1}{c}{Vehicle waiting time (s)}\\
\hline
Pre-defined & N/A & 1800 & 132713 \\
DQN & -15215 & 1252 & 77570 \\
DDQN & -15211 & 1246 & 76026  \\
PPO & -14570 & 1181 & 59677  \\
\hline\hline
\end{tabular}
\normalsize
\end{table}

Furthermore, it is worth comparing the training speed of RL by DQN, DDQN, and PPO on the same HPC cluster. The wall-clock times are recorded when finishing 600 episodes for each method. It takes DQN, DDQN, and PPO 6.5, 7.75, and 3.5 hours, respectively, to retrieve the optimal policies. PPO is faster than DQN and DDQN because it doesn't update deep neural networks at each step like DQN and DDQN do. Indeed, as an on-policy learning method, PPO updates the actor and critic networks after a certain number of steps once enough experiences are collected. Afterward, all the old experiences are discarded. In contrast, DQN and DDQN, which are off-policy learning methods, utilize the technique of experience replay to update Q-networks at each step.

In summary, learning-based intelligent traffic lights perform better than traffic lights with a predefined fixed-time plan. After comparing three RL methods, PPO, a policy-based method, is more efficient and effective than DQN and DDQN that are value-based methods in training intelligent traffic lights. 

\subsection{Traffic light phases with variable intervals}

The above study only considers the traffic light phases with fixed-time intervals as in most existing works \cite{Nishi2018}. Specifically, during the training, a selected traffic light phase stays for a constant interval of 10 seconds if it is different from the last phase. Although the current traffic light phase has a probability of being chosen and extended for another 5 seconds, there is not much flexibility in selecting various time intervals for the same traffic light phase. 

Here consider traffic light phases with variable time intervals, including 10, 15, 20, and 25 seconds for a chosen phase. Consequently, the agent needs to choose both a traffic light phase and an interval for the phase to stay during the action selection at each step. Therefore, instead of 8 available actions as illustrated in Fig.~\ref{fig:actions}, there are 32 available actions in this study, i.e., 32 combinations of traffic light phase and interval.

\begin{figure}
\centering{}
\includegraphics[scale=0.7]{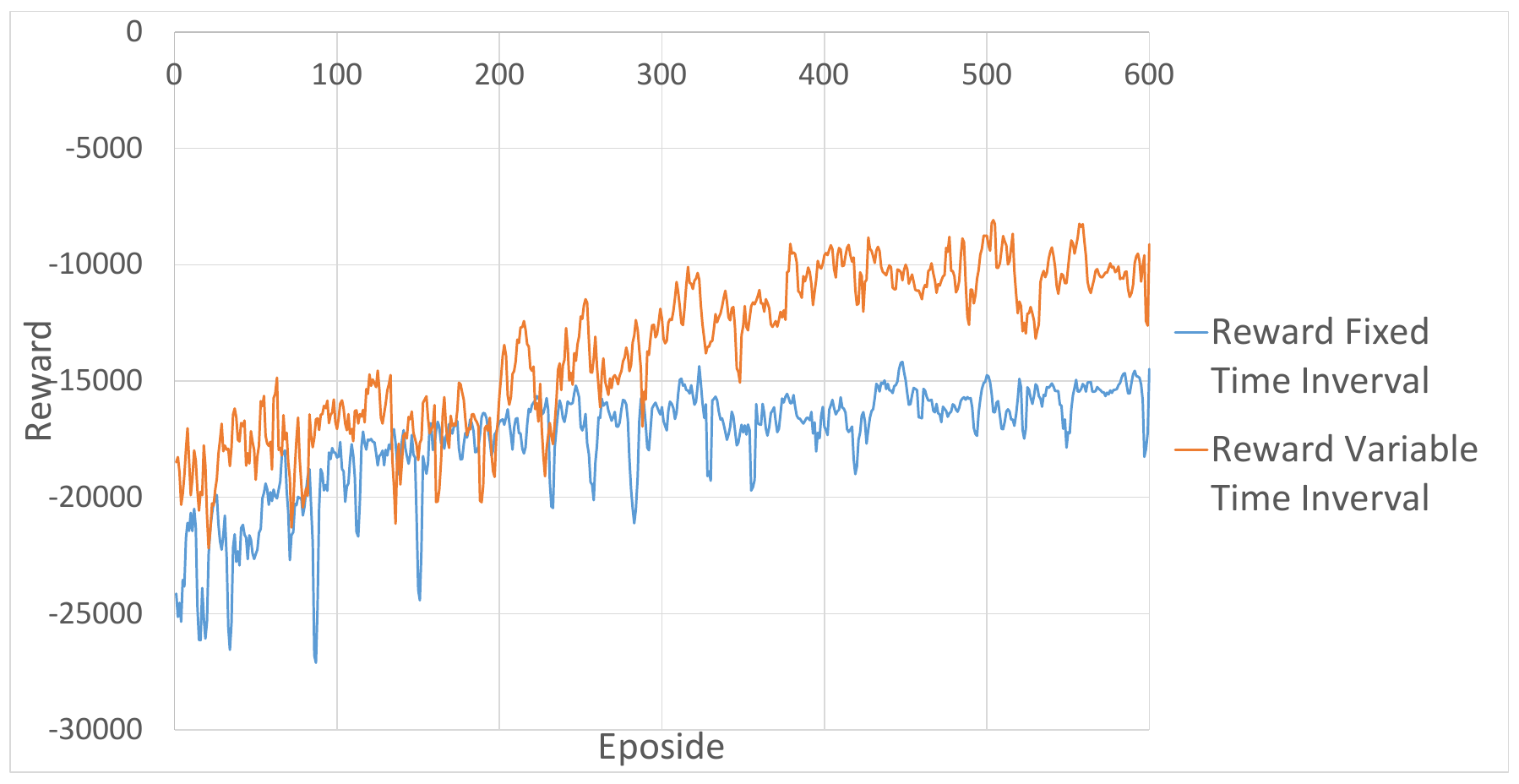}
\caption{Reward evolution when considering variable time intervals, compared to the one with fixed time intervals.}
\label{fig:RewardVar}
\end{figure}

Only the PPO method is used in this study because it has been shown to perform better than DQN and DDQN, and it needs less wall-clock time to converge. The actor and critic neural networks keep the same architectures as in Fig.~\ref{fig:Qnets}, except that the output layer of the actor network has 32 neurons instead of 8. Other parameters in training traffic lights via RL are the same as provided above. Fig.~\ref{fig:RewardVar} illustrates the reward evolution during the training via PPO when considering variable time intervals, compared to the one with fixed time intervals. It can be seen that considering traffic light phases with variable intervals results in a higher reward once converged. That means a better policy is achieved.

To quantitatively assess the induced policy when considering the traffic light phases with variable intervals, we conduct ten simulations under the obtained optimal policy and average the time to pass all vehicles and the vehicles' waiting time. The vehicles' passing time and total waiting time are 1076 seconds and 55028 seconds, respectively. Compared to the data resulting from the traffic light phases with fixed time intervals (the control agent was trained via PPO in the previous subsection), shown in Table~\ref{table:policyAssess}, the time reductions are 9$\%$ in the passing time and 8$\%$ in the total waiting time. We also calculate the average number of times the traffic light phases have been switched, i.e., the number of traffic light stops, under each optimal policy. We found that the number is dramatically reduced from 137 to 63 when considering variable time intervals. Although there are no significant improvements in the passing time and waiting time, much less number of action choosing results in a notable higher reward, as shown in Fig.~\ref{fig:RewardVar}.

\subsection{Environment and action disturbances} 

This study considers environment and action disturbances to demonstrate that learning-based intelligent traffic lights are robust. Two policies, named A and B, from previous PPO training are adopted. The difference is that Policy A is for the traffic light phases with variable time intervals, while policy B is for the light phases with fixed time intervals.

The environment disturbance is due to the occurrence of traffic collisions. We assume that there is a 2$\%$ probability with which a driver may not follow the traffic laws and cause the crash. Random number generations can introduce such an environment disturbance in the SUMO simulator. The collision can occur at a random location on any lane of either incoming or outgoing roads at an unexpected time. It is assumed that a traffic collision stays for 300 seconds, then the vehicles involved in the crash are removed and have no further impact on the traffic flow.

As utilized in the above, ten simulations with collisions occurring on the incoming roads or the outgoing road are conducted respectively to assess policies A and B, and the results are compared to the ones from predefined traffic lights, as listed in Table~\ref{table:environDis}. We separate the collision occurrence on the incoming and outgoing roads as two scenarios because collisions occurring on the incoming roads have a higher impact on the traffic flow than collisions occurring on the outgoing roads.

\begin{table}
\caption{Assessment data when collisions occur as environment disturbances}
\label{table:environDis}
\centering
\small
\renewcommand{\arraystretch}{1.25}
\begin{tabular}{c c c c}
\hline\hline
\multicolumn{1}{c} {Traffic light} & 
\multicolumn{1}{c}{Collision occurrence}  & 
\multicolumn{1}{c} {Vehicle passing time (s)} & 
\multicolumn{1}{c}{Vehicle waiting time (s)} \\
\hline
Predefined & incoming roads & 1820 & 135117 \\
Policy A & incoming roads & 1271 & 69884 \\
Policy B & incoming roads & 1260 & 70233 \\
Predefined & outgoing roads & 1800 & 132750 \\
Policy A & outgoing roads & 1190 & 64965 \\
Policy B & outgoing roads & 1192 & 66979 \\
\hline\hline
\end{tabular}
\normalsize
\end{table}

Table~\ref{table:environDis} shows that learning-based traffic lights perform much better than predefined traffic lights, the same as concluded from the above studies. Traffic lights under Policies A and B perform similarly with respect to the vehicles' passing time. In addition, Policy A is slightly better than policy B in reducing the vehicles' total waiting time. It shall be noted that the number of traffic light stops under policy A is much less than the one under policy B because variable-interval light phases are utilized in Policy A. An interesting insight in this study is that we retrain the traffic light by implementing the environment disturbances in training, but the new policy performs worse than Policies A and B, although it is better than predefined traffic lights.

Action disturbances can happen due to the malfunction of the traffic light. In this case (considering action disturbances only), after an action, i.e., selecting a light phase to switch or stay, is determined, the traffic light has a probability of 90 percent to switch to the desired light phase and a probability of 10 percent to switch to one of the other phases randomly. It is assumed that the environment is fully observable. Therefore, the traffic light phase vector, one of the state variables, is based on the actual traffic light phase instead of the one chosen by the agent.

In addition to Policies A and B, a new policy named Policy C is obtained via PPO, implementing action disturbance in SUMO for training. The assessment data of three policies, compared to the predefined traffic light, are listed in Table~\ref{table:actionDis}. Again, the learning-based traffic lights under Policies A, B, and C perform much better than the predefined traffic light, considering action disturbances due to traffic light malfunctions. Policy A is slightly better than Policy B; however, policy A results in fewer light stops than policy B. Although considering action disturbance in training, the induced policy (Policy C) doesn't operate the traffic light better than Policies A and B. However, Policy C may be useful in practice because it is from online training, while Policies A and B are induced from offline training.  

\begin{table}
\caption{Assessment data when action disturbances due to traffic light malfunction}
\label{table:actionDis}
\centering
\small
\renewcommand{\arraystretch}{1.25}
\begin{tabular}{c c c c}
\hline\hline
\multicolumn{1}{c} {Traffic light} & 
\multicolumn{1}{c} {Vehicle passing time (s)} & 
\multicolumn{1}{c}{Vehicle waiting time (s)} &
\multicolumn{1}{c}{number of light phase switches}  \\
\hline
Predefined & 1942 & 139863 & 129 \\
Policy A & 1187 & 58389 & 65 \\
Policy B & 1240 & 60932 & 132 \\
Policy C & 1348 & 76640 & 130 \\
\hline\hline
\end{tabular}
\normalsize
\end{table}

\begin{figure}
\centering{}
\includegraphics[scale=0.8]{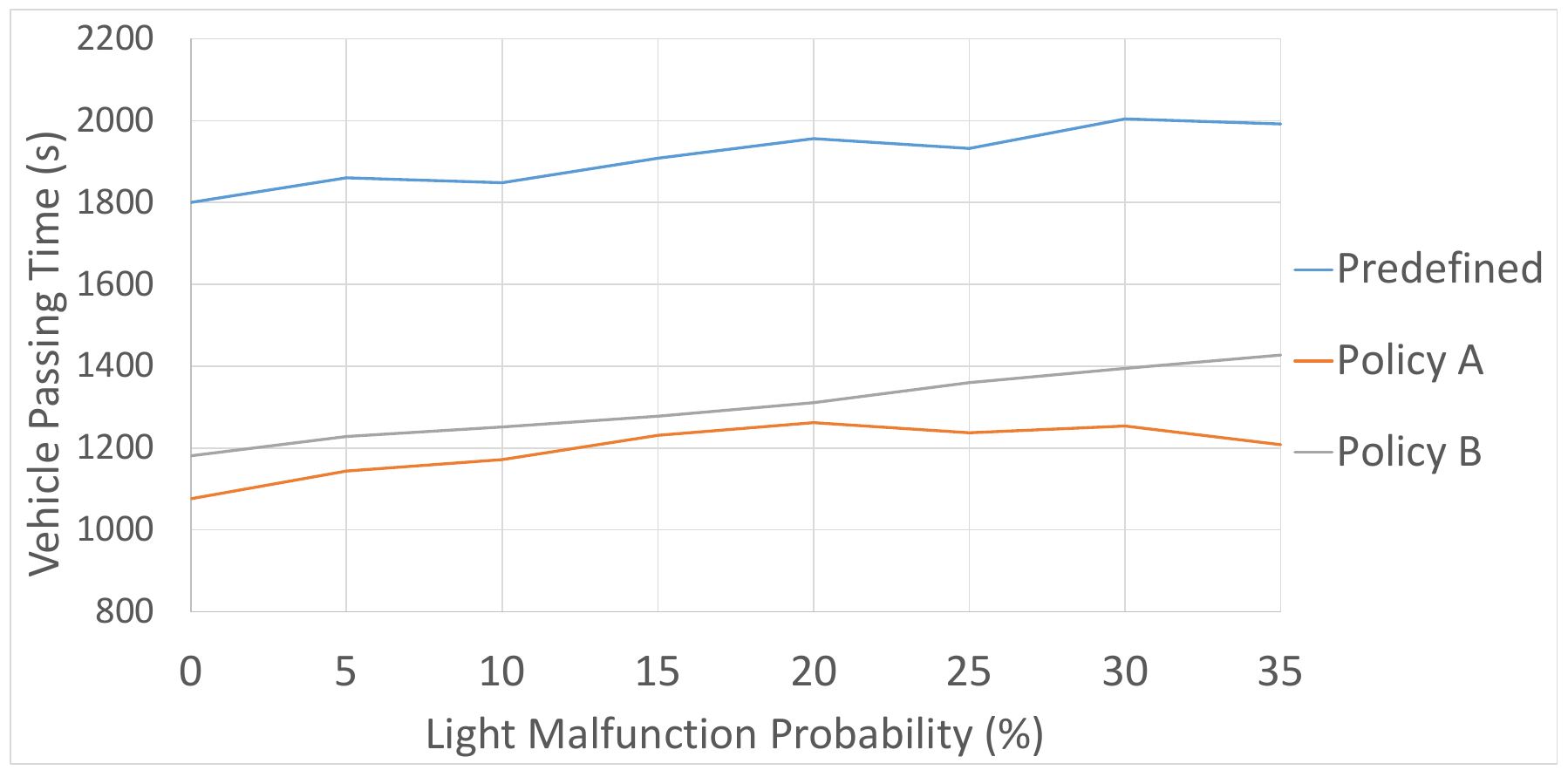}
\caption{Vehicle passing times considering various probabilities of traffic light malfunction.}
\label{fig:passingtime}
\end{figure}

In addition, We investigate the traffic light performances under the predefined plan, Policy A, and Policy B under various probabilities of action disturbances, i.e., considering different traffic light malfunction probabilities, from 5\% to 35\% at 5\% intervals. We conduct 50 simulations for every policy at each light malfunction probability and plot the averaged vehicle passing times in Fig.~\ref{fig:passingtime}. It can be seen that Policy A results in similar passing times with the traffic light malfunction probability up to 35\%, while the vehicle passing times under Policy B and the predefined plan gradually increase if the light malfunction gets worse. However, traffic lights under Policies A and B perform much better than the predefined traffic light, as concluded above. It shall be noted that based on our observation, when the light malfunction probability becomes larger than 35\%, the vehicle passing time under Policy A is notably increased as well.

\subsection{Unbalanced traffic flow} 
In the above studies, only balanced traffic flows are considered. In other words, each incoming road has the same traffic flow rates as indicated in Table~\ref{table:flowrate}. Here investigates the traffic light performance under Policy A in complex traffic flows, including balanced and unbalanced traffic, in which three scenarios are adopted continuously. The balanced flow rates, the same as in Table~\ref{table:flowrate}, are applied within the first 500 seconds. Then, the flow rates in the incoming east road are reduced by three-fourth for another 500 seconds while the flow rates keep the same in other incoming roads. At last, during the third 500 seconds, the flow rates in the incoming east road are back to normal while the other incoming roads' flow rates are reduced by three-fourth. 

For comparison, we utilize the predefined traffic light as used in the above studies. We also conduct another training to induce a new policy, named Policy D, directly from the aforementioned complex traffic flows. The averaged vehicle passing times and waiting times are compared in Table~\ref{table:Unbalanced}. Although Policy A is obtained from the learning with balanced traffic flows, the performance of traffic lights under this policy is moderately well, compared to the one under Policy D that is particularly learned from the complex traffic flows consisting of balanced and unbalanced traffic. Specifically, the vehicle passing time and waiting time from Policy D are only 8.3\% and 9.4\%, respectively, better than Policy A's. It is acceptable that policies learned from balanced traffic flows can also handle unbalanced traffic flows. However, when considering extremely unbalanced traffic flows, our other simulations \footnote{\url{https://github.com/YueZhu95/Intelligent-Traffic-Light-via-Reinforcement-Learning}} indicate that it would be better to learn the optimal policy directly from these scenarios.

\begin{table}
\caption{Assessment data of traffic lights under the predefined plan, Policy A, and Policy D in complex traffic flows consisting of balanced and unbalanced traffic.}
\label{table:Unbalanced}
\centering
\small
\renewcommand{\arraystretch}{1.25}
\begin{tabular}{c c c}
\hline\hline
\multicolumn{1}{c} {Traffic light} & 
\multicolumn{1}{c} {Vehicle passing time (s)} & 
\multicolumn{1}{c}{Vehicle waiting time (s)} \\
\hline
 Predefined & 4320 & 452401 \\
 Policy A & 3041 & 233483 \\
 Policy D & 2790 & 211528 \\
\hline\hline
\end{tabular}
\normalsize
\end{table}

\section{Conclusions and future works}
In this research, an intelligent traffic light learns to operate at a traffic intersection via RL properly. After comparing the performance of three DRL methods, DQN, DDQN, and PPO, and assessing their induced policies, PPO as a policy-based DRL method is better than value-based DRL methods such as DQN and DDQN. We consider various time intervals from which the control agent can choose for any light phase. Compared to the fixed-interval traffic lights that most existing works assumed, the traffic light with variable-interval phases generally can result in a shorter passing time, a less vehicles' waiting time, and a much smaller number of phase switches. We also study the scenarios in which there are environment disturbances due to collisions or action disturbances because of traffic light malfunction. Our simulations demonstrated that the optimal policies via offline training without disturbances were robust and performed well in those scenarios.  

This paper focuses on learning-based traffic light control at a single traffic intersection. Multiple traffic lights will be studied in the future, and multi-agent reinforcement learning (MARL) needs to be adopted. It shall be noted that a limited number of intervals are considered in this study so that the action space is still discrete. This work can be extended to constructing a continuous action space in which intervals within a time range are available. Consequently, some other policy-based methods, including Asynchronous Advantage Actor Critic (A3C) and Deep Deterministic Policy Gradient (DDPG) can be applied. In addition, both vehicles and pedestrians can be considered in a future work to design learning-based traffic lights.

\section{Data Availability Statement}

Some or all data, models, or code that support the findings of this study are available from the corresponding author upon reasonable request (list items).

%
%
\bibliography{ascexmpl-new}

\end{document}